\documentclass[11pt]{article}

\usepackage[preprint]{acl}

\usepackage{times}
\usepackage{latexsym} 
\usepackage{subcaption}
\usepackage{amsmath, amssymb, amsthm}
\usepackage{booktabs} 
\usepackage[T1]{fontenc}
\usepackage{amsfonts}
\usepackage[utf8]{inputenc}
\newcommand{\paratitle}[1]{\vspace{1.5ex}\noindent\textbf{#1}}
\newcommand{\inc}[1]{\textcolor{teal}{\scriptsize{(#1)}}}
\newcommand{\dec}[1]{\textcolor{red}{\scriptsize{(#1)}}}
\usepackage{microtype}

\usepackage{inconsolata}
\usepackage{multirow}
\usepackage{graphicx}
\usepackage{tcolorbox}

\usepackage{colortbl}
\definecolor{forward}{RGB}{84, 130, 53}
\definecolor{inverse}{RGB}{47, 85, 151}
\definecolor{resist}{RGB}{128, 0, 128}
\definecolor{rebound}{RGB}{133, 19, 33}

\definecolor{def}{RGB}{119, 228, 200}
\definecolor{thm}{RGB}{69, 53, 193}

\newtheorem{theorem}{Theorem}

\newtheorem{remark}{Remark}
\newtcolorbox{thmbox}[1][]{colback=thm!5!white,colframe=thm!60!black,boxsep=-4pt,grow to left by=4pt,left=10pt,grow to right by=4pt,right=10pt,top=10pt,bottom=10pt,#1}

\title{Entropy-Guided Token Dropout: Training Autoregressive Language Models with Limited Domain Data}

\author{
    \textbf{Jiapeng Wang}\textsuperscript{1}\thanks{\ \ Equal contribution.} \quad
    \textbf{Yiwen Hu}\textsuperscript{1}\footnotemark[1] \quad
    \textbf{Yanzipeng Gao}\textsuperscript{1} \quad
    \textbf{Haoyu Wang}\textsuperscript{1} \\ 
    \textbf{Shuo Wang}\textsuperscript{2}  \quad
    \textbf{Hongyu Lu}\textsuperscript{3} \quad
    \textbf{Jiaxin Mao}\textsuperscript{1} \quad
    \\
    \textbf{Wayne Xin Zhao}\textsuperscript{1}\thanks{\ \ Corresponding author.} \quad
    \textbf{Junyi Li}\textsuperscript{4}\footnotemark[2] \quad
    \textbf{Xiao Zhang}\textsuperscript{1}\footnotemark[2] \\
    \textsuperscript{1}Gaoling School of Artificial Intelligence, Renmin University of China. \\
    \textsuperscript{2}Tsinghua University.  
    \textsuperscript{3}WeChat, Tencent. \\
    \textsuperscript{4}Department of Data Science, City University of Hong Kong. \\
    {wangjp1010@ruc.edu.cn, batmanfly@gmail.com}
}

\begin{document}
\maketitle
\begin{abstract}
As access to high-quality, domain-specific data grows increasingly scarce, multi-epoch training has become a practical strategy for adapting large language models (LLMs). However, autoregressive models often suffer from performance degradation under repeated data exposure, where overfitting leads to a marked decline in model capability. 
Through empirical analysis, we trace this degradation to an imbalance in learning dynamics: predictable, low-entropy tokens are learned quickly and come to dominate optimization, while the model’s ability to generalize on high-entropy tokens deteriorates with continued training.  
To address this, we introduce \textbf{EntroDrop}, an entropy-guided token dropout method that functions as structured data regularization. EntroDrop selectively masks low-entropy tokens during training and employs a curriculum schedule to adjust regularization strength in alignment with training progress. 
Experiments across model scales from 0.6B to 8B parameters show that EntroDrop consistently outperforms standard regularization baselines and maintains robust performance throughout extended multi-epoch training. These findings underscore the importance of aligning regularization with token-level learning dynamics when training on limited data. Our approach offers a promising pathway toward more effective  adaptation of LLMs in data-constrained domains.

\end{abstract}

\section{Introduction}

The scaling of large language models (LLMs) has been the primary driver of recent advancements in artificial intelligence, with performance improvements largely correlated with increases in model size and training corpus volume~\cite{kaplan,Hoffmann}. However, this paradigm faces a fundamental bottleneck: the ``\emph{token crisis}''. The supply of high-quality human text available on the internet is finite and rapidly approaching depletion~\citep{Will}. This scarcity is particularly acute in high-value, specialized domains (e.g., mathematical proofs, legal briefs, complex agent trajectories)~\cite{Mitigating,Clinical,ling2025domain}, where expert-level data is inherently limited and cannot be easily scaled. Consequently, the field faces an urgent question: \emph{How can we continue advancing model capabilities when the supply of relevant token data is strictly limited?}

To mitigate data scarcity, a widely adopted and straightforward method is \emph{multi-epoch training}, in which models are repeatedly trained on the same dataset over multiple passes. 

Recent studies, however, indicate that autoregressive (AR) models are prone to ``\emph{multi-epoch training degradation}'' — repeated exposure to identical sequences tends to cause overfitting and superficial memorization, rather than meaningful generalization ~\citep{Repeat,Interpretability,Muennighoff23Scaling}.
In contrast to human learners, who deepen and refine their understanding through repeated exposure, autoregressive models tend to degenerate into pattern-specific memorization. When trained repeatedly on the same data, they often fail to learn robust, transferable representations of the underlying data distribution.

More recently, diffusion language models (DLMs)~\cite{ddpm,diffusionsurvey,llada} have emerged as a compelling alternative to AR models. Unlike AR approaches, DLMs are trained to reconstruct data from noise-corrupted inputs. By modeling the data distribution via an iterative denoising process, DLMs are essentially exposed to a substantially larger set of effective training signals derived from the same underlying corpus.  
Building on this insight, recent investigations have attempted to transplant this strategy into AR models. Notably, prior studies~\cite{superdatalearner,what2025} explored incorporating DLM-style input masking into AR architectures. They have shown that applying random input masking (i.e., token-level dropout) during multi-epoch training can induce data diversity and variation similar to those observed in DLMs. 

However, directly adopting this vanilla approach may entail significant limitations. First, uniformly dropping random tokens treats all tokens as equally informative, ignoring their distinct functional roles within a context. Intuitively, a more effective strategy would account for the varying impact of masking tokens of differing importance—for instance, by prioritizing tokens that are semantically or structurally critical. 
Second, applying uniform token dropout from the outset introduces strong input-level regularization that can disrupt early-stage optimization. This often leads to substantially slower initial convergence: models tend to underperform standard AR baselines in early epochs and require extended training to recover, thereby compromising overall training efficiency.

To address these limitations, we propose \textbf{Entropy-guided Token Dropout (EntroDrop)}, a principled approach designed to improve the utility of scarce data. Specifically tailored to the scarce subset rather than the entire corpus, our method is both adaptive and content-aware, employing two key techniques: 
(1) \textit{Entropy-guided targeting}: Instead of uniform random dropout, we prioritize masking low-entropy tokens—those predicted with high confidence across contexts. These tokens, which are accurately modeled early in training, offer diminishing returns with repeated exposure. By selectively masking them, we reduce redundant supervision in predictable regions and preserve high-entropy tokens that carry more informative gradients. This encourages the model to focus its learning capacity on uncertain, semantically rich content. 
(2) \textit{Curriculum-based scheduling}: We introduce a dynamic schedule that synchronizes with the model's learning trajectory. This ensures high training efficiency and lossless performance relative to standard baselines during the initial phase, while progressively intensifying regularization in later stages to effectively counteract overfitting. As a result, the model sustains informative learning for a longer duration and ultimately achieves superior performance.

Extensive experiments on models ranging from 0.6B to 8B parameters demonstrate the benefits of our approach. Results show that EntroDrop substantially prolongs the beneficial  training horizon on scarce data, delaying the performance collapse typically observed in standard multi-epoch training. On mathematical reasoning and code generation benchmarks, our method consistently outperforms strong regularization baselines, such as hidden dropout, weight decay and NEFTune, while preserving the model’s general capabilities. We view these results as an initial step toward more effective utilization of limited training data, and plan to further investigate the scalability of this approach under larger datasets and broader experimental settings.

\begin{figure*}[t]
    \centering
    \begin{subfigure}{0.32\textwidth}
        \centering
        \includegraphics[width=\linewidth]{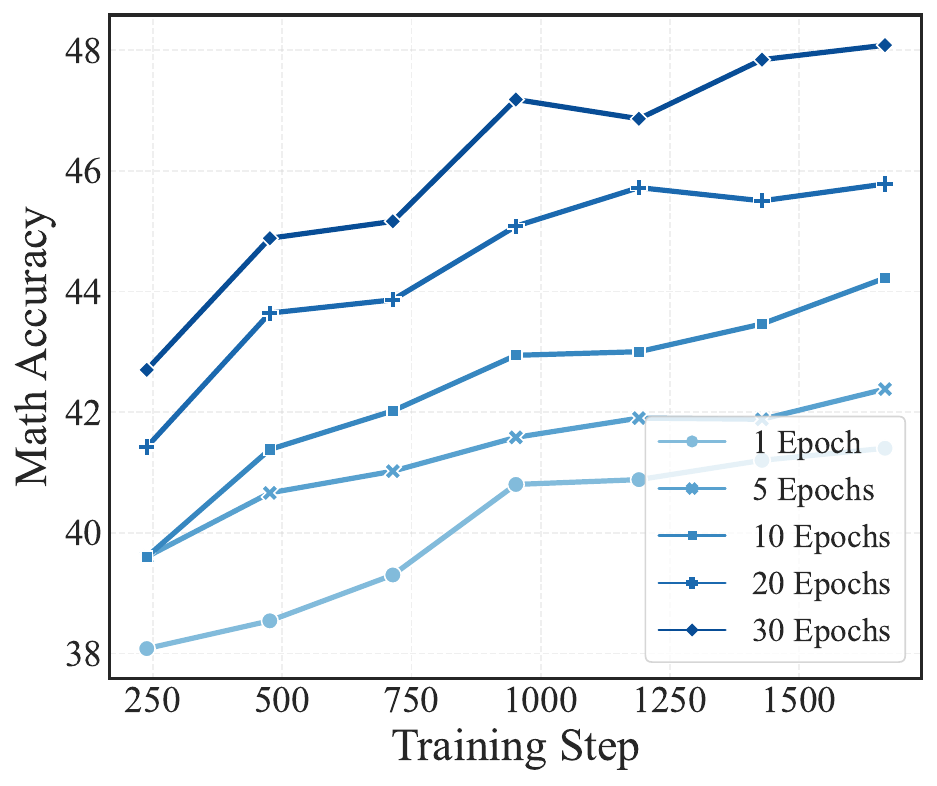}
        \caption{Performance Scaling w.r.t. Epochs}
        \label{fig:prelim_scaling}
    \end{subfigure}
    \hfill
    \begin{subfigure}{0.32\textwidth}
        \centering
        \includegraphics[width=\linewidth]{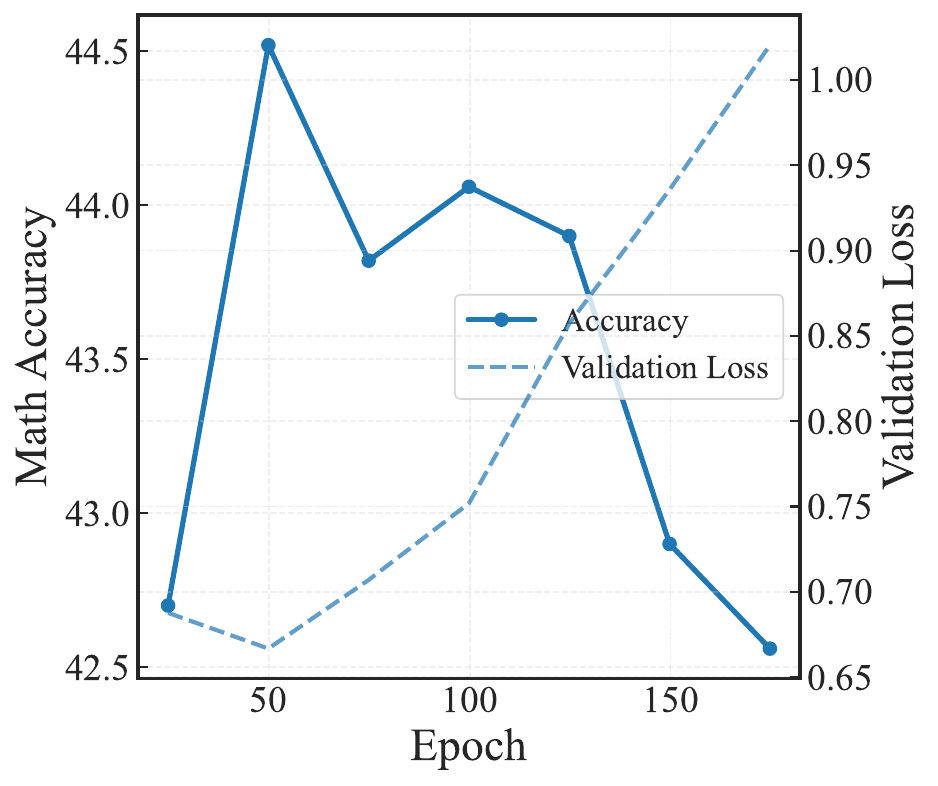}
        \caption{Accuracy vs. Validation Loss}
        \label{fig:prelim_loss}
    \end{subfigure}
    \hfill
    \begin{subfigure}{0.32\textwidth}
        \centering
        \includegraphics[width=\linewidth]{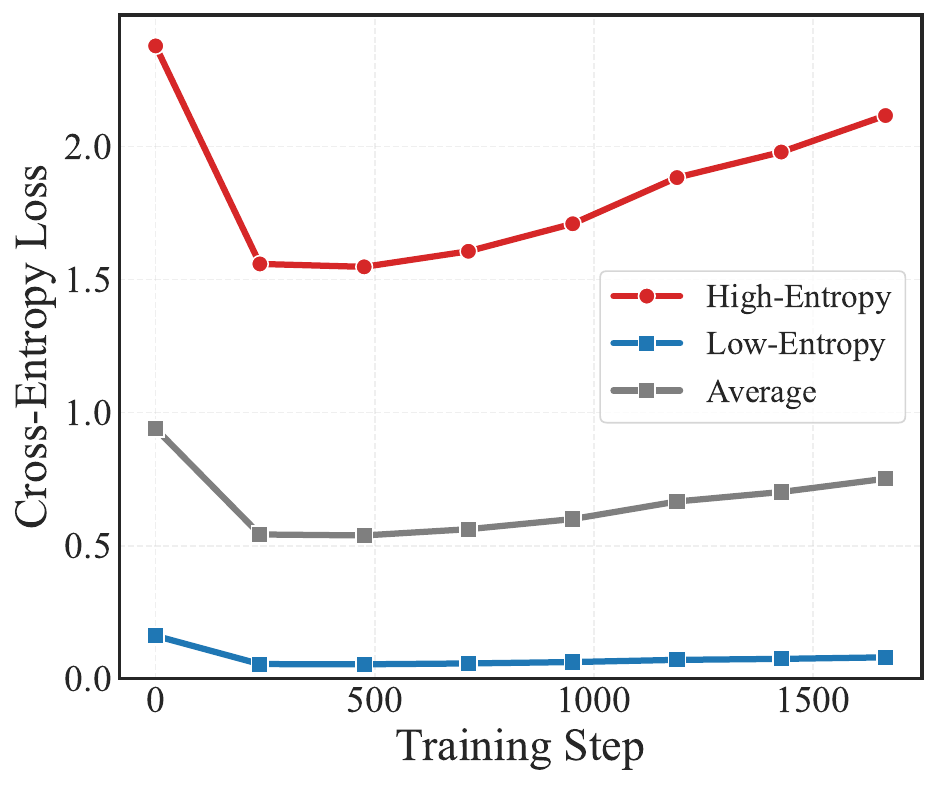}
        \caption{Loss Dynamics via Token Entropy}
        \label{fig:prelim_entropy}
    \end{subfigure}
    
    \caption{Preliminary analysis of learning dynamics under multi-epoch training with limited data.
    \textbf{(a)} Under a fixed compute budget, increasing the number of training epochs (up to a certain point) for domain-specific data \(D_{math}\) consistently improves accuracy, demonstrating that multi-epoch training is essential for data-constrained scenarios.
    \textbf{(b)} Excessive repetition eventually leads to performance degradation; accuracy declines as validation loss begins to rise.
    \textbf{(c)} A fine-grained analysis reveals that loss on low-entropy tokens remains stable and near zero, whereas loss on high-entropy tokens rebounds significantly after an initial decrease.}
    \label{fig:preliminary}
\end{figure*}

\section{Preliminary Study}
\label{ssec:preliminary}
In this section, we investigate the learning dynamics of LLMs when scaling limited domain-specific data through multi-epoch training under continual pre-training (CPT). We aim to empirically reconcile the trade-off between training sufficiency and overfitting, and subsequently analyze the mechanism of performance degradation using token entropy as a diagnostic signal.

\paragraph{RQ1: Is multi-epoch training beneficial for scarce data?}
We first examine whether repeatedly training on a limited corpus can yield performance gains before overfitting. 
To this end, we construct a controlled training corpus consisting of a small domain-specific dataset ($D_{math}$, 0.1B unique math tokens) and a substantially larger pool of general-domain data ($D_{general}$). Specifically, we adjust the volume of $D_{general}$ to maintain a total compute budget of 5B tokens, setting $|D_{general}| = 5\text{B} - (N \cdot |D_{math}|)$, where $N \in [1, 30]$ is the number of epochs for the math dataset. This design ensures that any observed performance variance is attributable to the frequency of domain-specific data exposure rather than the total number of training tokens.
Figure~\ref{fig:prelim_scaling} presents mathematical reasoning accuracy as a function of the repetition factor. We observe that increasing the repetition of $D_{math}$ consistently improves performance, with the model trained for 30 epochs significantly outperforming the single-epoch baseline. These results indicate that in data-constrained scenarios, single-pass training may provide insufficient learning signal, and that multi-epoch training can be an effective mechanism for improving domain performance.

\paragraph{RQ2: Does excessive repetition harm performance?}
While data repetition facilitates initial learning, extended training on the same corpus may also introduce performance degradation. Figure~\ref{fig:prelim_loss} reports mathematical accuracy and validation loss over a prolonged training horizon (up to 180 epochs). We observe that accuracy increases during early training, peaks at around 50 epochs, and subsequently declines sharply with continued repetition. This behavior illustrates a phenomenon we refer to as \textit{multi-epoch degradation}: although repetition is beneficial for extracting signal from limited data, excessive exposure can eventually push the model away from the generalization regime toward overfitting.

\paragraph{RQ3: What characterizes the degradation process?}
To investigate the mechanisms underlying performance degradation induced by excessive repetition, we adopt a fine-grained analysis that examines how learning dynamics differ across components of the input. As one such perspective, we analyze training behavior at the token level, distinguishing tokens by their information density. Specifically, we split the validation tokens into \textit{low-entropy} (top 50\% confidence) and \textit{high-entropy} (bottom 25\% confidence) groups based on the base model's likelihoods.
Figure~\ref{fig:prelim_entropy} shows a clear divergence in their loss trajectories. The loss on low-entropy tokens quickly drops and remains near zero, indicating stable and accurate modeling throughout training even under heavy repetition. Conversely, the loss on high-entropy tokens initially decreases, reaches a minimum and then rises markedly as training continues, reflecting a progressive deterioration in generalization.
This divergence suggests that multi-epoch degradation does not stem from a uniform decline in model performance, but rather from an imbalance in how learning capacity is allocated across tokens. Continued repetition primarily reinforces well-modeled, low-entropy regions of the data, yielding diminishing returns, while performance on high-entropy tokens degrades over time. These observations suggest that mitigating multi-epoch degradation requires regularization mechanisms that are selective rather than uniform, explicitly accounting for differences in token predictability and their distinct roles during training.

\section{Methodology}
\label{sec:method}

Building on the above analysis, we introduce \textbf{entropy-guided token dropout (EntroDrop)}, a structured regularization approach for multi-epoch training on scarce domain-specific data. The method is motivated by the observation that performance degradation under repeated training is highly non-uniform: overfitting concentrates on predictable, low-entropy tokens that rapidly saturate, while generalization on high-entropy tokens deteriorates.
Existing regularization strategies typically ignore this imbalance and apply uniform perturbations across tokens, leading to inefficient use of regularization capacity.
To address this issue, our approach incorporates two key techniques: (1) an entropy-based targeting mechanism that selectively regularizes predictable tokens, and (2) a curriculum schedule that balances early-stage learning with late-stage regularization.

\subsection{Formulation}
\label{subsec:formulation}

Formally, let \(\mathbf{X} = [\mathbf{x}_1, \dots, \mathbf{x}_T]\) denote the sequence of input embeddings for a training example drawn from the target-domain corpus. Token dropout is implemented by sampling a binary mask \(\mathbf{m} \in \{0,1\}^T\), where \(m_t = 0\) indicates that the token at position \(t\) is suppressed. 

We fill masked positions with a mask token embedding \(\bar{\mathbf{e}}\), yielding the regularized input:
\begin{equation}
\tilde{\mathbf{x}}_t = m_t \cdot \mathbf{x}_t + (1 - m_t) \cdot \bar{\mathbf{e}},
\label{eq:masked_input}
\end{equation}
where \(\bar{\mathbf{e}}\) is the static mean embedding of the vocabulary. The choice of \(\bar{\mathbf{e}}\) is discussed in Appendix~\ref{app:mask}.

At each training step \(j\), we sample a stochastic token dropout ratio \(\gamma_j \sim \mathcal{U}(0, \gamma_{\max}^{(j)})\), where the upper bound \(\gamma_{\max}^{(j)}\) is time-dependent and follows a curriculum schedule detailed in Section \ref{ssec:schedule}. Sampling $\gamma_j$ from a range of sparsity levels exposes the model to varying degrees of contextual corruption, promoting robustness to incomplete inputs while avoiding abrupt or overly aggressive regularization. Token dropout is applied exclusively to samples drawn from the scarce target-domain dataset \(\mathcal{D}_{{target}}\). All general-domain replay data remain unperturbed, ensuring stable background representations throughout training.

\subsection{Entropy-guided Token Targeting}
\label{ssec:entropy_guidance}

The core module of our method is entropy-guided token targeting, which conditions token dropout on token predictability. As shown in Section \ref{ssec:preliminary}, tokens within a corpus differ substantially in entropy. Low-entropy tokens, those predicted with high confidence, tend to converge rapidly to near-zero loss and remain well-modeled even under extensive repetition, contributing diminishing learning signal with continued optimization. In contrast, performance degradation under multi-epoch training is primarily associated with high-entropy tokens, whose loss increases as training progresses.

This asymmetry suggests that uniform random masking is poorly aligned with the observed degradation dynamics. Random perturbations may inefficiently regularize already saturated regions or unnecessarily corrupt high-entropy tokens that are critical for generalization. To address this mismatch, we bias token dropout toward low-entropy tokens, selectively suppressing redundant supervision while preserving informative gradients on less predictable regions.

Concretely, we compute the contextual entropy \(H(x_t)\) of each token using the base model and treat it as a proxy for token predictability. A binary targeting gate \(g_t\) is then computed as:
\begin{equation}\label{eq:entropy_mask}
g_t = \mathbb{I} \left( H(x_t) \le \operatorname{Percentile}_k\bigl( H(\mathbf{X}) \bigr) \right),
\end{equation}
where \(k\) specifies the fraction of lowest-entropy tokens eligible for regularization (e.g., the bottom 50\%). The final masking decision is sampled conditionally as follows:
\begin{equation}
\label{eq:entropy_mask:prob}
P(m_t = 0) = \gamma_j \cdot g_t,
\end{equation}
such that only tokens falling within the low-entropy subset are subject to dropout, and the overall intensity is controlled by the dropout ratio \(\gamma_j\).

By selectively masking predictable tokens, the model is trained on perturbed yet semantically coherent contexts, which discourages over-specialization on memorized patterns. At the same time, supervision on uncertain and decision-critical tokens is preserved, enabling learning capacity to be directed toward regions where generalization remains more challenging.
From a computational perspective, entropy-guided targeting introduces minimal overhead. In the multi-epoch setting, entropy is computed once per sample and reused across epochs, amortizing the additional cost to less than $1/N$ of the total training budget for $N$ epochs.

\subsection{Curriculum-based Dropout Schedule}
\label{ssec:schedule}

While targeted token dropout can mitigate redundancy, its effectiveness depends critically on both timing and intensity. Applying strong regularization too early in training can interfere with representation learning and substantially slow convergence. Conversely, weak or delayed regularization in later stages may be insufficient to prevent the onset of memorization.

To reconcile this trade-off, we introduce a curriculum schedule that synchronizes dropout intensity with the model’s learning dynamics. The maximum allowable dropout rate at training step \(j\) is defined as
\begin{equation}\label{eq:gamma_schedule}
\gamma_{\max}^{(j)} = \frac{\gamma_{\max}}{1 + \exp\big(-k \cdot (j - j_0)\big)} ,
\end{equation}
where \(\gamma_{\max}\) denotes the upper bound of the dropout ratio, \(k\) controls the sharpness of the transition, and \(j_0\) represents the characteristic training step at which regularization begins to dominate. In practice, \(j_0\) is chosen to coincide with the regime where the baseline transitions from productive learning to memorization-prone behavior, as estimated from a validation curve. This schedule ensures that early training proceeds in a near-lossless regime, enabling efficient adaptation to the target domain, while regularization is progressively intensified when pronounced risk of memorization.

\subsection{Discussion}
Entropy-guided token targeting and curriculum scheduling constitute a structured regularization approach for multi-epoch training on scarce data. By adapting both \emph{where} and \emph{when} regularization is applied, the method aims to better align regularization with observed training dynamics, enabling autoregressive language models to make more effective use of limited corpora under repeated exposure.

To clarify the optimization mechanism of token-level regularization, we conduct theoretical analysis from the perspective of training gradients. A key finding reveals that with proper configuration of the masking rate and targeting parameters, EntroDrop effectively reduces training gradient variance, laying a theoretical foundation for stable model training.
Detailed proof and analyses can be found in the Appendix~\ref{sec:theo_analysis}.

\begin{thmbox}
\begin{theorem}[Gradient Variance Bound]
\label{thm:TokenDropout:Gradientbound}
Under assumptions listed in Table~\ref{tab:assumption}, the gradient variance (w.r.t. model parameters $\theta$) of the EntroDrop loss $L^{\mathcal{M}}$ can be bounded by that of the loss $L$ without token dropout:
\begin{equation*}
\begin{aligned}
\mathbb{V}(\nabla_{\theta}L^{\mathcal{M}}(\theta;x)) &\le \mathbb{V}(\nabla_{\theta}L(\theta;x)) \cdot (1 - \gamma_j \cdot \alpha) \\
&\hphantom{{}={}} + G^2 \cdot (\sigma + \delta)^2 \cdot (\gamma_j \cdot \alpha)^2,  
\end{aligned}
\end{equation*}
where $\gamma_j$ is the curriculum mask rate in Eq.~\eqref{eq:entropy_mask:prob}, $\alpha: = \sum_{t=1}^T g_t / T$ denotes the proportion of low-entropy tokens (with $T$ as the input sequence length) governed by the percentile in $g_t$ in Eq.~\eqref{eq:entropy_mask}, and $G, \sigma, \delta$ are model and data parameters related to gradient Lipschitz continuity and embedding properties (see Table~\ref{tab:assumption}).
\end{theorem}
\end{thmbox}
From the upper bound in Theorem~\ref{thm:TokenDropout:Gradientbound},  when the curriculum mask rate satisfies: 
$$\gamma_j \leq \mathbb{V}(\nabla_{\theta}L(\theta;x)) ~\big/~ [G \cdot (\sigma + \delta) \sqrt{\alpha}]^2,$$
we can obtain 
$$\mathbb{V}(\nabla_{\theta}L^{\mathcal{M}}(\theta;x)) \le \mathbb{V}(\nabla_{\theta}L(\theta;x)), $$
which implies that EntroDrop can achieve a lower gradient variance during training compared to the variance of the original (unmasked) gradient. This result demonstrates the effectiveness of the proposed token dropout in reducing gradient variance.

\begin{remark}[\emph{Token Dropout Supplements Data-Level Regularization}]
 
\emph{
Theorem~\ref{thm:TokenDropout:Gradientbound} also reveals that token dropout in our EntroDrop acts as an implicit regularizer by attenuating gradient variance~\citep{pmlr-v134-haochen21a}, which helps suppress updates along high-curvature directions associated with overfitting~\citep{jelassi2022towards, wu2023implicit}.  
Existing regularization methods operate at the \emph{model level}, achieving anti-overfitting by constraining parameters, optimizing network structures, or adjusting training logic (e.g., \citep{wu2021r, shen2020powernorm,NEFTune}).
In contrast, our EntroDrop works at the \emph{data level}: it achieves regularization effects by directly processing input data (e.g., token-level manipulation), without altering the model’s architecture or training paradigm.
}

\end{remark}

\begin{figure*}[t]
    \centering
    \begin{subfigure}{0.32\textwidth}
        \centering
        \includegraphics[width=\linewidth]{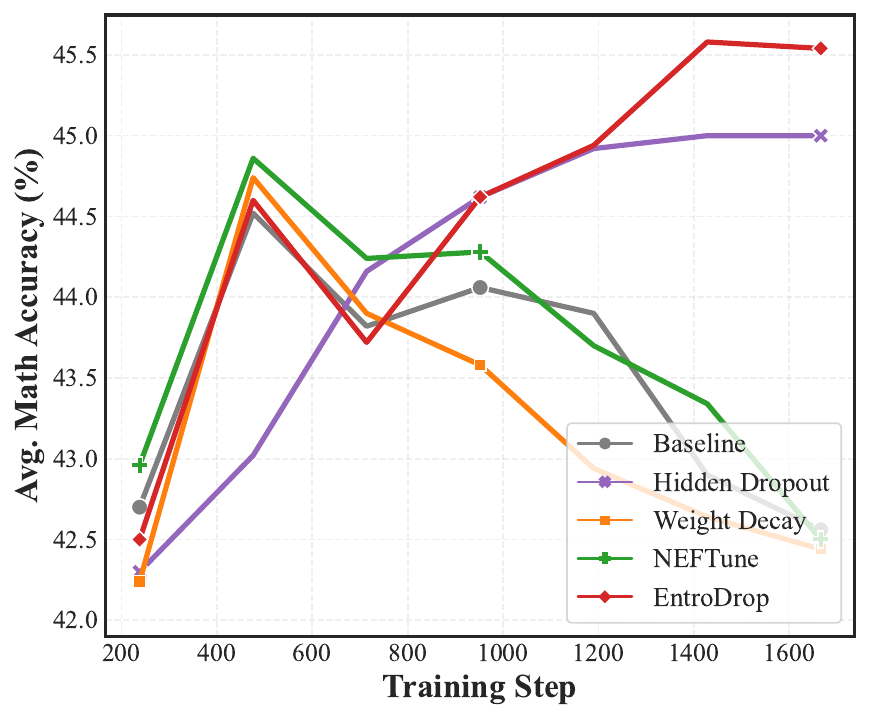}
        \caption{Effectiveness of Token Dropout}
        \label{fig:sub_comparison}
    \end{subfigure}
    \hfill
    \begin{subfigure}{0.32\textwidth}
        \centering
        \includegraphics[width=\linewidth]{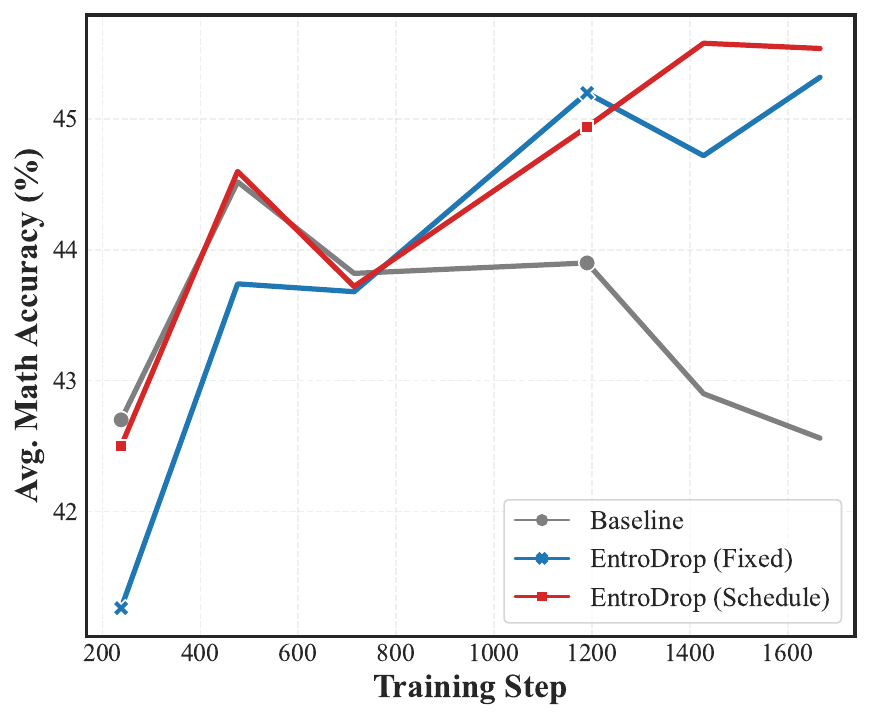}
        \caption{Impact of Masking Schedule}
        \label{fig:sub_schedule}
    \end{subfigure}
    \hfill
    \begin{subfigure}{0.32\textwidth}
        \centering
        \includegraphics[width=\linewidth]{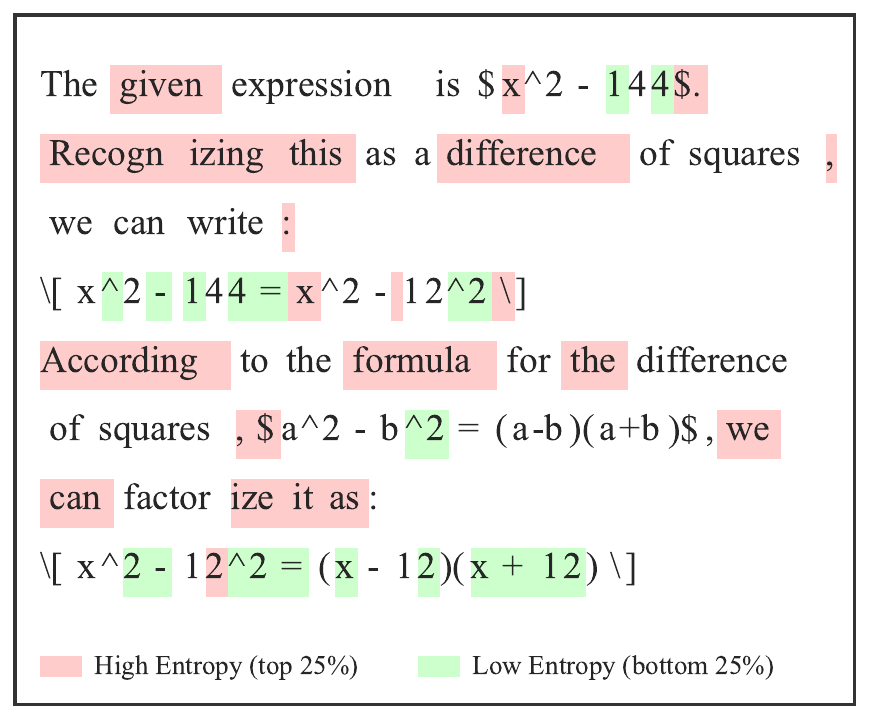}
        \caption{Entropy Visualization}
        \label{fig:sub_entropy}
    \end{subfigure}
    
    \caption{
    Analysis of learning dynamics. 
    \textbf{(a)} Comparison with standard regularization: Our method (EntroDrop) significantly extends the effective training duration compared to baselines (e.g., NEFTune, weight decay) on Qwen3-0.6B.
    \textbf{(b)} Impact of masking ratio schedule: The dynamic curriculum schedule matches the baseline's early learning efficiency while preventing collapse in later stages, outperforming a fixed ratio.
    \textbf{(c)} A representative mathematical reasoning example illustrating that high-entropy tokens often correspond to pivotal semantic steps, while low-entropy tokens represent predictable patterns.
    }
    \label{fig:ablation_schedule}
\end{figure*}

\begin{remark}[\emph{Gradient Variance Reduction Enhances Generalization}]
\emph{
The gradient variance bound in Theorem~\ref{thm:TokenDropout:Gradientbound} supports EntroDrop's generalization advantage: reducing gradient variance alleviates overfitting during training by mitigating high-curvature updates~\cite{neu2021information}. As shown in Figure~\ref{fig:ablation_schedule}, as training steps increase, baselines degrade rapidly, while EntroDrop enhances data utilization via sustained performance growth.
}
\end{remark}

\section{Experiments}
\subsection{Experimental Setup}

\paratitle{Models and Datasets.}
We evaluate our method based on several open-weight LLMs of varying scales and architectures, including {Llama3.1-8B-Instruct}~\cite{llama3}, {Qwen3-0.6B} and {Qwen3-1.7B}~\cite{qwen3}, on both math reasoning and code generation tasks.
The math reasoning adopts a suite of standard benchmarks, including {SVAMP}~\cite{svamp}, {GSM8K}~\cite{gsm8k}, {MATH}~\cite{math}, {CollegeMath}~\cite{college-math}, and {OlympiadBench}~\cite{OlympiadBench}. For code generation, we evaluate on {HumanEval}~\cite{humaneval}, {MBPP}~\cite{mbpp}, and {LiveCodeBench V1}~\cite{livecodebench}. We also evaluate the models on a suite of out-of-domain benchmarks: {HellaSwag}~\cite{hellaswag} and {PIQA}~\cite{piqa} (commonsense reasoning), {ARC}~\cite{arc} (scientific reasoning), {MMLU}~\cite{mmlu} (general knowledge), and {IFEval}~\cite{ifeval} (instruction following) with preserving the model's fundamental general capabilities. We use OpenCompass~\cite{2023opencompass} for evaluation.

\paratitle{Baselines.}
We compare our method against several regularization baselines: hidden dropout~\cite{Dropout}, weight decay~\cite{Decoupled}, and NEFTune~\cite{NEFTune}.
For in-depth analysis and ablation studies, we conduct experiments by default on {Qwen3-0.6B}. Detailed hyperparameters and implementation configurations are provided in Appendix \ref{app:setup}.

\paratitle{Training Settings.}
Our experiments are conducted under a continual pre-training (CPT) setting. For the 0.6B models, training is performed with a total budget of 3B tokens, while the 1.7B and 8B models are trained with a budget of 1B tokens.
Training data consist of a mixture of target-domain and general-domain corpora with a fixed mixing ratio of 6:4. The target-domain corpus contains 0.02B unique tokens and is therefore reused across multiple epochs to satisfy the allocated token budget, resulting in repeated exposure over multiple training passes. The general-domain corpus is sufficiently large and is sampled with negligible repetition.
\begin{table*}[t]
\small
\centering
\caption{Main performance comparison on mathematical reasoning and general capability benchmarks. For a fair comparison, results are reported from the checkpoint with the highest domain-average score for each method. Our method consistently achieves superior accuracy in mathematical reasoning while effectively preserving or even enhancing general capabilities.}
\label{tab:main_results}

\resizebox{\textwidth}{!}{
\begin{tabular}{l  cccccl  cccccl  l}
\toprule
\multirow{2}{*}{\textbf{Method}} &
\multicolumn{6}{c}{\textbf{Mathematical Reasoning}} &
\multicolumn{6}{c}{\textbf{General Capabilities}} &
\multirow{2}{*}{\textbf{Total Avg.}} \\
\cmidrule(lr){2-7} \cmidrule(lr){8-13}
& \textbf{SVAMP} & \textbf{GSM8K} & \textbf{MATH} & \textbf{College} & \textbf{Olym.} & \textbf{Avg.}
& \textbf{Hella} & \textbf{ARC-C} & \textbf{PIQA} & \textbf{MMLU} & \textbf{IFEval} & \textbf{Avg.}
& \\
\midrule
\multicolumn{14}{c}{\textit{\textbf{Backbone: Qwen3-0.6B}}} \\
\midrule

baseline & 78.9 & 60.4 & 40.6 & 29.5 & 13.2 & 44.52 & 45.0 & 56.6 & 68.3 & 50.0 & 44.3 & \underline{52.84} & \underline{48.68} \\

weight decay &
78.7 & 60.3 & 40.5 & 29.7 & 14.5 &
44.74 \inc{+0.49\%} &
44.9 & 55.9 & 68.1 & 49.9 & 43.2 &
52.40 \dec{-0.83\%} &
48.57 \dec{-0.23\%} \\

hidden dropout &
80.7 & 62.5 & 40.2 & 28.1 & 13.5 & \underline{45.00} \inc{+1.08\%} &
43.9 & 60.3 & 68.8 & 48.4 & 35.5 &
51.38 \dec{-2.76\%} &
48.19 \dec{-1.01\%} \\

NEFTune &
79.5 & 60.4 & 41.1 & 29.1 & 14.2 &
44.86 \inc{+0.76\%} &
45.0 & 56.3 & 67.7 & 50.0 & 42.7 &
52.34 \dec{-0.95\%} &
48.60 \dec{-0.16\%} \\

vanilla token dropout & 79.6 & 63.0 & 38.3 & 29.1 & 13.9 & 44.78 \inc{+0.58\%} &
45.3 & 55.6 & 68.4 & 49.2 & 43.6 & 52.42 \dec{-0.79\%} &
48.60 \dec{-0.16\%}
\\

\midrule
\textbf{EntroDrop} &
81.3 & 63.0 & 38.9 & 29.1 & 15.4 &
\textbf{45.54} \inc{+2.29\%} &
45.3 & 60.7 & 68.5 & 49.7 & 42.5 &
\textbf{53.34} \inc{+0.95\%} &
\textbf{49.44} \inc{+1.56\%} \\

\midrule
\multicolumn{14}{c}
{\textit{\textbf{Backbone: Qwen3-1.7B}}} \\
\midrule

baseline &
86.3 & 74.4 & 50.4 & 36.7 & 22.5 &
54.06 &
58.4 & 77.6 & 74.0 & 62.8 & 50.7 &
64.69 &
59.38 \\

weight decay &
86.7 & 75.5 & 50.6 & 36.6 & 23.0 &
\underline{54.48} \inc{+0.78\%} &
58.4 & 76.6 & 74.0 & 62.6 & 52.3 &
64.79 \inc{+0.15\%} &
59.64 \inc{+0.44\%} \\

hidden dropout &
87.1 & 75.8 & 48.0 & 36.1 & 23.3 &
54.06 \inc{+0.00\%} &
58.3 & 78.0 & 74.3 & 62.5 & 51.5 &
64.96 \inc{+0.42\%} &
59.51 \inc{+0.22\%} \\

NEFTune &
85.9 & 73.2 & 51.4 & 37.2 & 23.7 &
54.28 \inc{+0.41\%} &
58.4 & 77.6 & 74.3 & 62.7 & 52.9 &
\textbf{65.17} \inc{+0.74\%} &
\underline{59.73} \inc{+0.59\%} \\

vanilla token dropout & 86.6 & 74.8 & 51.1 & 34.8 & 23.0 &54.06 \inc{+0.00\%}&
58.2 & 77.9 & 73.6 & 62.5 & 52.7 & 64.98 \inc{+0.45\%} &
59.52 \inc{+0.23\%}
\\

\midrule
\textbf{EntroDrop} &
86.5 & 75.4 & 51.9 & 36.4 & 24.9 &
\textbf{55.02} \inc{+1.78\%} &
58.3 & 77.0 & 74.3 & 62.7 & 53.2 &
\underline{65.09} \inc{+0.62\%} &
\textbf{60.06} \inc{+1.15\%} \\

\midrule
\multicolumn{14}{c}{\textit{\textbf{Backbone: Llama3.1-8B-Instruct}}} \\
\midrule

baseline &
86.2 & 77.3 & 41.4 & 31.6 & 16.7 &
50.64 &
73.1 & 87.5 & 79.9 & 67.9 & 69.8 &
\textbf{75.60} &
\underline{63.14} \\

weight decay &
84.8 & 81.3 & 45.3 & 31.5 & 15.0 &
\underline{51.58} \inc{+1.86\%} &
73.1 & 85.7 & 80.4 & 66.9 & 66.8 &
74.58 \dec{-1.35\%} &
63.08 \dec{-0.10\%} \\

hidden dropout &
86.2 & 79.4 & 42.4 & 31.7 & 17.3 &
51.40 \inc{+1.50\%} &
73.4 & 85.4 & 80.4 & 67.1 & 67.4 &
74.74 \dec{-1.14\%} &
63.07 \dec{-0.11\%} \\

NEFTune &
87.0 & 81.0 & 42.5 & 29.7 & 14.2 &
50.88 \inc{+0.47\%} &
73.1 & 85.1 & 81.0 & 67.9 & 66.4 &
74.70 \dec{-1.19\%} &
62.79 \dec{-0.55\%} \\

vanilla token dropout & 85.3 & 79.2 & 44.3 & 31.3 & 15.6 & 51.14 \inc{+0.99\%}&
72.6 & 85.1 & 80.5 & 68.2 &65.8 & 74.44 \dec{-1.53\%}&
62.79 \dec{-0.55\%}
\\

\midrule
\textbf{EntroDrop} &
86.2 & 80.2 & 44.2 & 32.1 & 16.1 &
\textbf{51.76} \inc{+2.21\%} &
73.1 & 86.1 & 80.9 & 67.6 & 69.7 &
\underline{75.48} \dec{-0.16\%} &
\textbf{63.62} \inc{+0.76\%} \\

\bottomrule

\bottomrule
\end{tabular}
}
\end{table*}

\begin{table*}[h]
\centering
\caption{Main performance comparison on code generation and general capability benchmarks using Qwen-0.6B.}
\label{tab:code_general}
\resizebox{\textwidth}{!}{%
\begin{tabular}{l  cccl  cccccl  l}
\toprule
\multirow{2}{*}{\textbf{Method}} &
\multicolumn{4}{c}{\textbf{Code Generation}} &
\multicolumn{6}{c}{\textbf{General Capabilities}} &
\multirow{2}{*}{\textbf{Total Avg.}} \\
\cmidrule(lr){2-5} \cmidrule(lr){6-11}
& \textbf{HumanEval} & \textbf{MBPP} & \textbf{LiveCodeBench V1} & \textbf{Avg.}
& \textbf{Hella} & \textbf{ARC-C} & \textbf{PIQA} & \textbf{MMLU} & \textbf{IFEval} & \textbf{Avg.} & \\
\midrule
baseline       
& 33.5 & 30.8 & 22.7 & 29.00
& 45.1 & 59.0 & 68.4 & 49.8 & 42.4 & 52.94
& 43.96 \\

weight decay   
& 37.2 & 29.6 & 25.2 & 30.67 \inc{+5.76\%}
& 45.4 & 62.4 & 68.2 & 49.8 & 42.4 & \textbf{53.64} \inc{+1.32\%}
& \underline{45.03} \inc{+2.43\%} \\

hidden dropout 
& 37.8 & 31.2 & 23.0 & {30.67} \inc{+5.76\%}
& 44.4 & 57.6 & 68.2 & 48.9 & 13.5 & 46.52 \dec{-12.13\%}
& 40.58 \dec{-7.69\%} \\

NEFTune        
& 34.1 & 29.2 & 26.3 & 29.87 \inc{+3.00\%}
& 44.6 & 55.9 & 67.6 & 49.8 & 42.7 & 52.12 \dec{-1.55\%}
& 43.78 \dec{-0.41\%} \\

vanilla token dropout &
37.8&29.0&31.9  & \underline{32.90}\inc{+13.45\%} &
44.4 & 56.6 & 67.8 & 48.8 &43.3 & 52.18 \dec{-1.44\%} &
44.95 \inc{+2.25\%}

\\

\midrule
\textbf{EntroDrop}           
& 39.6 & 28.8 & 30.5 & \textbf{32.97} \inc{+13.69\%}
& 44.5 & 58.0 & 68.1 & 49.0 & 45.1 & \underline{52.94} \inc{+0.00\%}
& \textbf{45.45} \inc{+3.39\%} \\

\bottomrule
\end{tabular}%
}
\end{table*}

\subsection{Main Results}

For a fair comparison, Table~\ref{tab:main_results} reports results from the checkpoint with the highest domain-average score. 
All regularization methods achieve higher performance than the baseline in the target mathematical domain, and EntroDrop consistently yields the largest gains across all model backbones. For instance, on Qwen3-0.6B, EntroDrop improves the math domain average by +2.29\% over the baseline,  substantially outperforming weight decay (+0.49\%) and NEFTune (+0.76\%).

Figure~\ref{fig:sub_comparison} illustrates the training dynamics of the mathematical benchmark score on Qwen3-0.6B. The baseline peaks early and then undergoes rapid performance degradation. Standard methods such as weight decay and NEFTune fail to mitigate this decline. In contrast, both hidden dropout and EntroDrop significantly extend the effective training window: performance continues to rise steadily, even after the baseline collapses, and ultimately surpasses the baseline’s peak, enabling more thorough and stable learning.

Another crucial observation is the preservation of general capabilities. While standard regularization, including hidden dropout, often leads to performance degradation or catastrophic forgetting on general benchmarks, EntroDrop largely maintains, and in some cases slightly improves, general-domain performance relative to the baseline. By preserving general domain tokens and pivotal semantic anchors, EntroDrop safeguards the model’s core reasoning and linguistic structures during domain adaptation.

Table~\ref{tab:code_general} extends these findings to code generation. On Qwen3-0.6B, EntroDrop achieves an 11.62\% relative improvement, confirming that the benefits of entropy-guided regularization generalize beyond mathematical reasoning to other structured domains.

\subsection{Ablation Analysis}

\paratitle{Ablation on Entropy Targets.}
We investigate the impact of targeting tokens with different information densities. 
Figure \ref{fig:sub_entropy} visualizes token-level entropy on a representative mathematical reasoning example. We observe a clear structural separation between high- and low-entropy tokens. We observe high-entropy tokens tend to correspond to pivotal semantic elements that determine the global reasoning trajectory. In contrast, low-entropy tokens are predominantly associated with local syntactic realizations and intermediate derivations.
The ablation results in Table \ref{tab:ablation_entropy} further substantiate the role of information density in effective regularization. 
Targeting high-entropy tokens yields marginally improvement over uniform random masking.
In contrast, selectively masking low-entropy tokens produces the strongest performance improvements. This result aligns with our token-level analysis: low-entropy tokens are already accurately modeled early in training and contribute little additional learning signal under repetition. Regularizing these regions reduces redundant optimization pressure while preserving informative gradients on high-entropy tokens, thereby mitigating the imbalance in training dynamics. 

\begin{table}[h]
\centering
\caption{Ablation study on entropy-based masking targets. Masking low-entropy tokens yields the most significant performance gain, confirming that regularizing redundant, well-modeled regions is more effective
}
\label{tab:ablation_entropy}
\resizebox{1.0\linewidth}{!}{%
\begin{tabular}{lcc}
\toprule
\textbf{Targeting Strategy} & \textbf{Masking Scope} & \textbf{Avg. Math Acc} \\ \midrule
Random Masking & All Tokens & 44.78 \\ \midrule
High-Entropy (Top 50\%) & Pivotal Info & 44.84 (+0.06) \\
Low-Entropy (Bottom 50\%) & Easy Patterns & \textbf{45.32} (+0.55) \\ \bottomrule
\end{tabular}%
}
\end{table}

\paratitle{Ablation on Ratio Schedule.}
We further analyze the importance of the dynamic masking schedule by comparing it with a static alternative, where the masking ratio $\gamma$ is fixed at its maximum value $\gamma_{\text{max}}$ from the beginning of training.  
As shown in Figure~\ref{fig:sub_schedule}, applying strong masking uniformly from the outset impedes early learning: the model exhibits slower initial convergence and consistently underperforms the baseline in early epochs.  
In contrast, our curriculum-based schedule begins with a low masking ratio, allowing the model to learn efficiently during the initial phase, matching the baseline’s early progress. As training proceeds and overfitting risks rise, the gradually increasing masking intensity helps sustain learning without collapse, enabling continued performance gains in later epochs where the baseline stagnates or declines.  
This ablation confirms that a gradual, learning-aware increase in regularization strength effectively balances early training efficiency with long-term robustness.

\subsection{Impact of Dropout Ratio Strategies}
\label{ssec:ratio}
We investigate the sensitivity of our model to the maximum masking proportion $\gamma_{max}$ in the dynamic sampling strategy. To comprehensively evaluate the training dynamics, we report two diagnostic metrics alongside the best average benchmark accuracy:
(1) Best epoch: The training epoch of target data where the validation loss reaches its global minimum. This serves as a proxy for the onset of overfitting.
(2) Min validation loss: The lowest validation loss achieved during training. A lower value implies a better theoretical fit to the underlying data distribution.
The results are summarized in Table~\ref{tab:ablation_ratio}. Increasing the masking ratio $\gamma_{max}$ significantly delays the onset of overfitting (best epoch) and improves theoretical generalization (min val loss). However, a moderate ratio yields the best downstream accuracy.
It is worth noting that token dropout does not eliminate overfitting entirely; rather, it postpones its occurrence. For instance, with $\gamma_{max}=0.1$, validation loss begins to rise after 50 epochs, whereas the baseline overfits much earlier.

\begin{table}[h]
\centering
\caption{Sensitivity analysis of the maximum masking ratio ($\gamma_{max}$). While increasing $\gamma_{max}$ significantly postpones the onset of overfitting and leads to lower theoretical loss, a moderate ratio ($\gamma_{max}=0.1$) achieves the optimal balance for downstream task performance.} 
\label{tab:ablation_ratio}
\resizebox{1.0\linewidth}{!}{%
\begin{tabular}{lccc}
\toprule
 & \textbf{Avg. Math Acc} & \textbf{Best Epoch} & \textbf{Min Val Loss} \\ \midrule
Baseline ($\gamma=0$) & 44.3 & 25 & 0.704 \\ \midrule
$\gamma_{max}=0.05$ & 44.86 & 50 & 0.692 \\
$\gamma_{max}=0.1$ & \textbf{44.94} & 50 & 0.689 \\
$\gamma_{max}=0.2$ & 43.80 & 100 & 0.680 \\
$\gamma_{max}=0.4$ & 42.68 & \textbf{$>$125} & \textbf{0.670} \\ \bottomrule
\end{tabular}%
}
\end{table}

\section{Related Work}
\subsection{Multi-epoch Training for LLM}
The rapid depletion of high-quality public text, referred to as the ``token crisis'', has made data efficiency a critical research focus~\cite{Will}. As the acquisition of new unique tokens becomes increasingly difficult, multi-epoch training has become a prevalent practice, particularly through the upsampling of high-quality data. However, the specific effects of this approach remain under-investigated, and there is a notable lack of discussion on how to optimize such repetitive training. In fact, recent studies show that this approach yields diminishing returns for AR LLM models. Unlike human learning, where repetition consolidates understanding, AR models suffer from degradation when trained for too many epochs, leading to overfitting and a loss of generalization capabilities~\cite{Muennighoff23Scaling,Repeat}.
Scaling laws for data-constrained regimes further quantify this limitation, suggesting that the benefits of extra epochs plateau quickly compared to adding fresh data~\cite{Muennighoff23Scaling,Interpretability,yan2025larger}. While synthetic data generation is a potential alternative, it often requires complex engineering pipelines and strict quality control to avoid model collapse~\cite{Synthetic,team2025kimik2}. Consequently, our work introduces a systematic investigation and a regularized training paradigm that effectively mitigates overfitting and deeply exploits the intrinsic value of the data.

\subsection{Data Efficiency and Regularization Techniques}
Regularization is essential for preventing overfitting, particularly when data is limited. Traditional techniques, such as hidden state dropout~\cite{Dropout}, word dropout~\cite{Regularizing,MaxMatch,gi-drop} and weight decay~\cite{Decoupled}, are standard in deep learning. However, in the context of LLMs, some of these traditional techniques often prove insufficient to prevent the severe overfitting during multi-epoch training~\cite{Repeat}. Recent advances in diffusion models demonstrate superior data efficiency compared to AR language models under data-constrained regimes~\cite{DiffusionBeats,superdatalearner,what2025}. This has motivated efforts to adapt similar principles to AR models, notably through token dropout (i.e., input masking), which introduces stochasticity at the token level to discourage memorization and promote robust feature learning.
Our method builds on this idea with entropy-guided token dropout: tokens are dropped selectively based on model confidence, targeting low-entropy (easily memorized) regions while preserving semantically rich content. This yields more precise and effective regularization.

\section{Conclusion}
This paper examines the challenge of repetitive data training in LLMs and introduces entropy-guided token dropout as a strategy to enhance the utility of scarce data. We show that performance decay during repeated exposure possibly stems from a learning imbalance: models overfit to predictable, low-entropy tokens while losing generalization on high-entropy, informative segments. By selectively masking tokens according to their predictability and adopting a curriculum-based schedule, our approach aims to improve how scarce data is learned, enabling models to acquire deeper and more transferable knowledge. Experimental results across multiple benchmarks demonstrate that this targeted regularization delays overfitting and sustains stable training over longer cycles. These findings offer a promising direction for addressing the “token crisis,” indicating that effective scaling can be achieved not only through more data, but more efficient and adaptive data utilization. 

\section*{Limitations}

While our evaluation is primarily focused on domains like mathematics and code, extending this framework to a broader range of tasks will further characterize its generalizability. We validated our method on models up to 8B parameters, its efficacy when scaling to much larger datasets and model sizes remains to be explored. Finally, while pre-computed entropy serves as an efficient proxy for predictability, future research could explore using dynamic training loss as a real-time signal to drive token dropout, potentially offering even more adaptive regularization as the model evolves.


\bibliography{custom}

\appendix

\section{Experimental Settings}
\label{app:setup}
\subsection{Implementation Details}
We train using Megatron-LM~\cite{megatron} with a sequence length of 4,096, a global batch size of 512, and a peak learning rate of 1e-5 with cosine decay. For mathematical reasoning, we use the NuminaMath~\cite{numina_math_datasets} dataset; for general-domain data, we use Infinity-Instruct~\cite{li2025infinityinstructscalinginstruction}; and for code generation, we use rStar-Coder~\cite{liu2025rstarcoderscalingcompetitivecode} dataset.
\subsection{Hyper-parameter Tuning}
\begin{figure}[h]  
\centering
\includegraphics[width=0.5\textwidth]{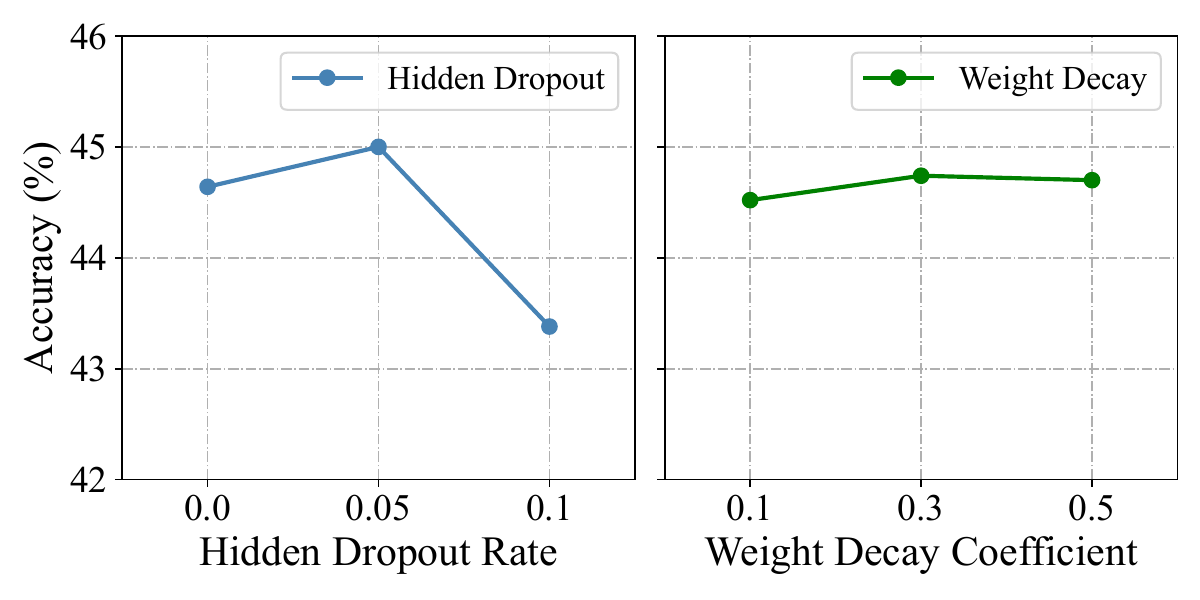}
\caption{Hyper-parameter tuning for baselines.}
\label{fig:appendix}
\end{figure}

As shown in Table~\ref{tab:ablation_ratio}, our method uses $\gamma_{\max} = 0.1$. For baseline regularization methods, we perform hyperparameter tuning: for hidden dropout, we test dropout rates of 0.05, and 0.1; for weight decay, we test coefficients of 0.1, 0.3, and 0.5. For NEFTune, we adopt the recommended hyperparameter $\alpha=5$ to scale the uniform noise injected into the embeddings by a factor of $\alpha/\sqrt{Ld}$. Results are summarized in Figure~\ref{fig:appendix}, and the best-performing configuration for each method is reported in the main results table.

\subsection{Choice of Mask Token Embedding}
\label{app:mask}
A natural implementation is to use the zero vector as \(\bar{\mathbf{e}}\). However, we observed that this leads to frequent gradient explosion (e.g., infinite gradients) during continual pretraining. To mitigate this, we adopt a mean interpolation strategy and set \(\bar{\mathbf{e}}\) to the average of the base model’s token embedding matrix:
\[
\bar{\mathbf{e}} = \frac{1}{|\mathcal{V}|} \sum_{w \in \mathcal{V}} \mathbf{E}[w],
\]
where \(\mathcal{V}\) is the vocabulary and \(\mathbf{E}[w] \in \mathbb{R}^d\) denotes the embedding of token \(w\). Empirically, this simple choice significantly stabilizes training and eliminating gradient divergence, without requiring additional parameters or fine-tuning.
\section{Comparison of Loss Dynamics}
\begin{figure}[h]  
\centering
\includegraphics[width=0.4\textwidth]{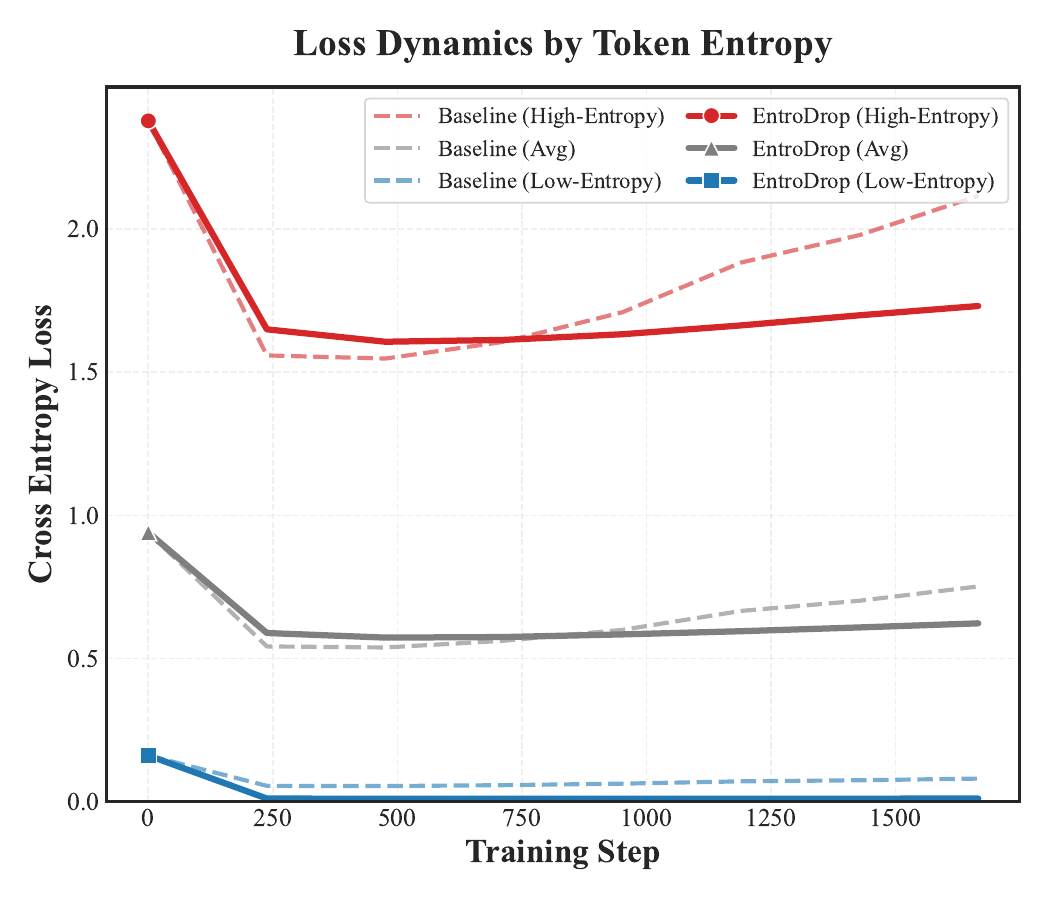}
\caption{Comparison of Loss Dynamics by Token Entropy.}
\label{fig:loss}
\end{figure}

To further examine how our method mitigates multi-epoch degradation, we compare the cross-entropy loss dynamics categorized by token entropy against the baseline (Figure \ref{fig:loss}). Our method effectively suppresses the rebound of validation loss, maintaining a lower and more stable loss trajectory.

\section{Theoretical Analysis}
\label{sec:theo_analysis}
In this section, we analyze the effectiveness of token dropout from the perspective of variance reduction.
Our main result indicates that, once the values of $\gamma_j$ and $g_t$ in section~\ref{ssec:entropy_guidance} are properly set, token dropout effectively reduces the variance of the training gradients, thereby leading to more stable model training.

Before proceeding with the formal analysis, we remark that our analysis is built on several assumptions, which are listed in detail in Table~\ref{tab:assumption}.
As we can see, Assumptions A1-2 are largely regarded as empirically plausible. 
While Assumptions A3-4 are aligned with our proposed method, we highlight that Assumptions A5 and A6 impose stronger restrictions on the robustness of LLMs.
Under the above assumptions, we prove Theorem~\ref{thm:TokenDropout:Gradientbound} as follows.

\begin{table*}[b]
\centering
\caption{Model and Data Assumptions}
\label{tab:assumption}
\begin{tabular}{lp{3.5cm}p{10.5cm}}
\toprule
\textbf{ID} & \textbf{Name} & \textbf{Definition} \\ \midrule
A1 & Bounded Variance & The standard deviation of low-entropy embeddings is bounded: $\sigma = \sqrt{\mathbb{V}(E[x_i]_{i \in S})} < +\infty$. \\ \hline
A2 & Embedding Offset & The global average embedding $\bar{e} = \frac{1}{|V|}\sum_{w \in V} E[w]$ and low-entropy token expectations satisfy: $\|\bar{e} - \mathbb{E}_{x \in \mathcal{D}_{dom}}[E[x_i]]\|_2 \le \delta$ for $i \in S$. \\ \hline
A3 & Mask Independence & Mask events $\{m_i\}$ are mutually independent: $P(m_i=1 \mid m_j=1) = P(m_i=1), \forall i \neq j$. \\ \hline
A4 & Partial Masking & Masking probability $P(m_i=1) = \gamma_j \cdot g_i \in [0, \Gamma]$, where $\gamma_j$ is the curriculum rate and $g_i$ is the gating function. \\ \hline
A5 & Gradient Lipschitz & The gradient $\nabla_\theta L$ is $G$-Lipschitz continuous w.r.t. low-entropy embeddings: $\|\nabla_\theta L(\theta;x, e_1) - \nabla_\theta L(\theta;x, e_2)\|_2 \le G \cdot \|e_1 - e_2\|_2$. \\ \hline
A6 & Gradient Independence & The gradients induced by token-wise loss $\nabla \log p_\theta(x_i)$ are independent with each other. \\
\bottomrule
\end{tabular}
\end{table*}


\begin{proof}[Proof of Theorem~\ref{thm:TokenDropout:Gradientbound}]
First, utilize the law of total variance to decompose the masked gradient variance:
\begin{equation*}
\begin{aligned}
\mathbb{V}(\nabla_\theta L^{\mathcal{M}}(\theta)) &= \mathbb{E}_m[\mathbb{V}(\nabla_\theta L(\theta) \mid m)]\\ 
&+\mathbb{V}_m[\mathbb{E}(\nabla_\theta L(\theta) \mid m)].   
\end{aligned}
\end{equation*}
We then bound the expected conditional variance term $\mathbb{E}_m[\mathbb{V}(\nabla_\theta L(\theta) \mid m)]$.
By A6, gradients of tokens are independent in an AR model. Let $S$ be the set of low-entropy tokens ($|S| = \alpha T$). 
For $i \in S$, $P(m_i=1) = \gamma_j$. If masked, the embedding is constant, so $\mathbb{V}(\nabla_\theta l_i \mid m_i=1) = 0$. 
Summing over the sequence we get
\begin{equation*}
\begin{aligned}
&\mathbb{E}_m[\mathbb{V}(\nabla_\theta L(\theta) \mid m)]\\ =\ &\frac{\alpha (1 - \gamma_j)}{T}\mathbb{V}_S(\theta) + \frac{1-\alpha}{T}\mathbb{V}_{\bar{S}}(\theta). 
\end{aligned}
\end{equation*}
Notice that the original variance is 
\begin{equation*}
\mathbb{V}(\nabla_\theta L(\theta)) = \frac{1}{T}[\alpha \mathbb{V}_S(\theta) + (1-\alpha)\mathbb{V}_{\bar{S}}(\theta)]
\end{equation*}
We have
\begin{equation}\label{eq:expect_variance}
\mathbb{E}_m[\mathbb{V}(\nabla_\theta L(\theta) \mid m)] \le \mathbb{V}(\nabla_\theta L(\theta)) \cdot (1 - \gamma_j \alpha).
\end{equation}
Then, we bound the Variance term of the Conditional Expectation.
Using the Lipschitz continuity of the gradient (A5), bounded variance (A1), and the embedding bounds (A2, A6), we obtain
\begin{equation*}
\|\mathbb{E}(\nabla_\theta L \mid m) - \mathbb{E}(\nabla_\theta L)\|_2 \le G \cdot (\sigma + \delta) \cdot \frac{1}{T}\sum_{i \in S} m_i.
\end{equation*}
Applying the variance definition 
\begin{equation*}
\mathbb{V}_m[\mathbb{E}(\nabla_\theta L \mid m)] \le G^2 (\sigma + \delta)^2 \mathbb{E}_m \left[ \left( \frac{1}{T} \sum_{i \in S} m_i \right)^2 \right].
\end{equation*}
For large $T$, the second moment of the Binomial distribution $\sum m_i \sim \mathrm{Bin}(\alpha T, \gamma_j)$ (A3, A4) yields 
\begin{equation*}
\begin{aligned}
&\lim_{T\to\infty}\mathbb{E}\left[\left(\frac{1}{T}\sum m_i \right)^2\right] \\
=& \lim_{T\to\infty}\left\{\left[\mathbb{E} \left[\frac{1}{T}\sum m_i \right]\right]^2 + \mathbb{V}\left[\frac{1}{T}\sum m_i\right]\right\} \\ 
=& (\gamma_j \alpha)^2 + \lim_{T\to\infty}\frac{\gamma_j \alpha(1-\gamma_j)}{T} \\
=& (\gamma_j \alpha)^2.
\end{aligned}
\end{equation*}
Thus,
\begin{equation}\label{eq:variance_expectation}
\mathbb{V}_m[\mathbb{E}(\nabla_\theta L \mid m)] \le G^2 (\sigma + \delta)^2 (\gamma_j \alpha)^2.
\end{equation}
Combining Eq.~(\ref{eq:expect_variance}) and Eq.~(\ref{eq:variance_expectation})  yields the final result.
\end{proof}

\paragraph{Remark}
It is easy to see that the right side of Theorem 1 is a quadratic function of $\gamma_j \cdot \alpha$.
By taking the derivative with respect to the masking rate $\gamma_{j}\cdot\alpha$, the theoretical optimal masking rate can be derived as
\begin{equation}\label{eq:optimal_schedule}
\gamma_{j}\cdot\alpha = \frac{\mathbb{V}(\nabla_{\theta}L(\theta;x))}{2G^{2}\cdot(\sigma+\delta)^{2}}.
\end{equation}

Although $G, \sigma, \delta$ exist in theory, they are often intractable to compute experimentally \citep{qistable}, so is the optimal masking rate consequently.
However, Eq.~(\ref{eq:optimal_schedule}) rigorously guarantees the existence and uniqueness of the optimal mask schedule.
Empirically, we observe that $\gamma_{\max}=0.1, k=50\%$ effectively achieves the desired outcome, whereas alternative choices fail to produce superior results, which aligns with our theoretical analysis.

\end{document}